\documentclass[journal]{IEEEtran}

\ifCLASSINFOpdf
\else
\fi
\usepackage{amsmath}
\usepackage{algorithmic}

\usepackage{array}

\usepackage{stfloats}
\usepackage{url}

\usepackage{times}  
\usepackage{helvet} 
\usepackage{courier}  
\usepackage{graphicx}  
\usepackage[square,sort,comma,numbers]{natbib}  
\usepackage{caption} 
\usepackage{amsmath}
\usepackage{multirow}
\usepackage{textcomp,booktabs}
\usepackage{colortbl}
\definecolor{mygray}{gray}{0.9}
\usepackage{diagbox}
\usepackage{marvosym}

\begin{document}

%
\title{Multi-Classifier Interactive Learning for Ambiguous Speech Emotion Recognition}
%
%
%

%
\author{\IEEEauthorblockN{Ying Zhou, Xuefeng Liang \textsuperscript{(\Letter)}, Yu Gu, Yifei Yin, Longshan Yao }\\
	\IEEEauthorblockA{\textit{School of Artificial Intelligence, Xidian University, China} \\ xliang@xidian.edu.cn}\thanks{This work has been submitted to the IEEE for possible publication. Copyright may be transferred without notice, after which this version may no longer be accessible.}}

\bibliographystyle{IEEEtran}

\maketitle

\begin{abstract}
In recent years, speech emotion recognition technology is of great significance in industrial applications such as call centers, social robots and health care. The combination of speech recognition and speech emotion recognition can improve the feedback efficiency and the quality of service. Thus, the speech emotion recognition has been attracted much attention in both industry and academic. Since emotions existing in an entire utterance may have varied probabilities, speech emotion is likely to be ambiguous, which poses great challenges to recognition tasks. However, previous studies commonly assigned a single-label or multi-label to each utterance in certain. Therefore, their algorithms result in low accuracies because of the inappropriate representation. Inspired by the optimally interacting theory, we address the ambiguous speech emotions by proposing a novel multi-classifier interactive learning (MCIL) method. In MCIL, multiple different classifiers first mimic several individuals, who have inconsistent cognitions of ambiguous emotions, and construct new ambiguous labels (the emotion probability distribution). Then, they are retrained with the new labels to interact with their cognitions. This procedure enables each classifier to learn better representations of ambiguous data from others, and further improves the recognition ability. The experiments on three benchmark corpora (MAS, IEMOCAP, and FAU-AIBO) demonstrate that MCIL does not only improve each classifier's performance, but also raises their recognition consistency from moderate to substantial.
\end{abstract}

\begin{IEEEkeywords}
Speech emotion recognition, Interactive learning, Multi-classifier approach.
\end{IEEEkeywords}

%

\section{Introduction}



 \IEEEPARstart{W}{ith} the rapid development of Artificial Intelligence, the technology of speech emotion recognition (SER) is becoming deeply involved into a wide bank of industrial applications, e.g. call center, health care, social robot, to name a few. The previous research reported that, in the daily communication of human-being, language accounts for 45\%, of which text accounting for only 7\% while emotional expression accounting for 38\%. Thus, emotion plays an important role in human communications. A high-quality SER does not only make human machine interaction more naturally, but also improve the efficiency and effectiveness of industrial services. For instance, NTT DoCoMo enables the intelligent customer service system to detect whether customers express a negative emotion. If yes, the phone call will be switched to the human service. Sony AIBO pet robot is able to recognize a few speech emotions, and adjusts its personality and interactive behavior according to the owner's emotion. In addition, SER technology is being applied in tutorial systems to improve social interaction abilities of children with autism spectrum disorders.

\begin{figure}[t!]	
	\begin{center}
		\includegraphics[width=6.5cm]{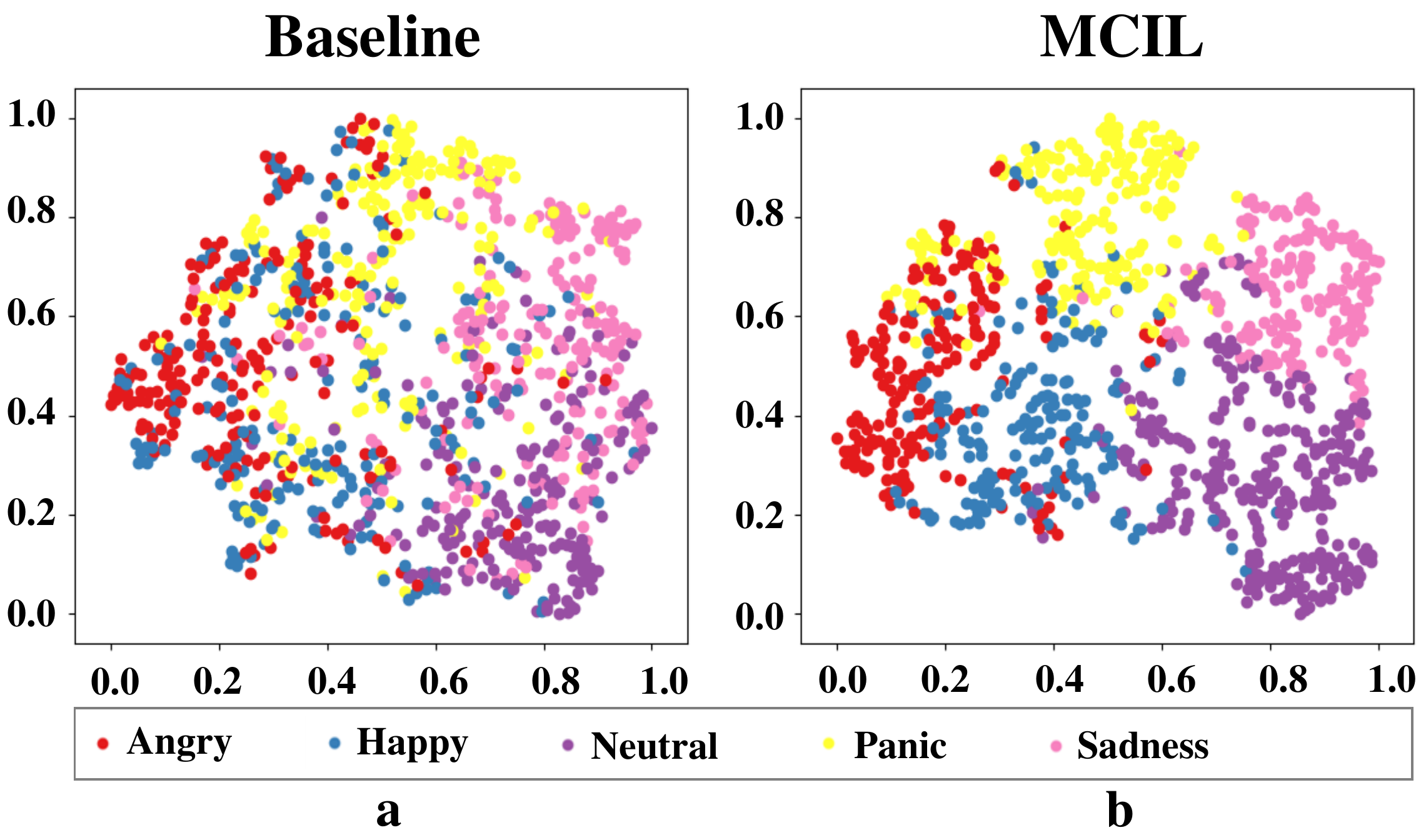}
	\end{center}
	\vspace{-.1in}
	\caption{Feature visualization of five emotions learned by VGG (a) before and (b) after employing our MCIL on Mandarin Affective Speech corpus. Clearly, MCIL enables VGG to learn a better representation of ambiguous data.}
	\label{fig:tsnevisualization}
\end{figure}

However, the performance of current SER technologies remains suboptimal due to the ambiguity of emotions. The psychology study \cite{ortony1990cognitive} demonstrated that speech emotions were somewhat ambiguous, which was also confirmed by the research using a statistical model \cite{tao2009multiple}. From the perspective of machine learning, we found the classification boundaries between emotion categories are not clear as well, as shown in Fig.\ref{fig:tsnevisualization}.a. Nevertheless, the conventional studies \cite{trigeorgis2016adieu,ma2018emotion} often assumed that emotions could be distinguishable, hence, assigned a precise/single label to each of them, which may not represent speech emotions very well. Commonly, they resulted in low accuracies.

Recently, a few studies tried to model the emotion ambiguity in their methods. For example, Fayek \textit{et al.} \cite{fayek2016modeling} designed a soft-target label. Lotfian \textit{et al.} \cite{lotfian2018predicting} considered multiple emotions in one utterance as a multi-task problem. Ando \textit{et al.} \cite{ando2019speech} used multi-label to represent the emotion ambiguity. Multi-label means each sample is assigned to a set of target labels, in which every label is certain, but emotions existing in utterances usually have varied probabilities. In addition, due to all these methods heavily relying on the statistics of experts’ voting, they have limited generalization to be applied to the databases without voting information, such as Mandarin Affective Speech (MAS) dataset. More interesting, we have observed that SER methods performed inconsistently on these emotional categories \cite{dai2019learning, ando2019speech}. The possible reason is that these methods are good at learning features of some specific emotions respectively. To our best knowledge, there is no one approach that can integrate the strengths of these methods into one.

When faced with this complicated issue of inconsistent recognition of ambiguous data, human beings have a wise strategy against it. Bahrami \textit{et al.} \cite{bahrami2010optimally} found that interpersonal communication can improve a person's ability to disambiguate uncertain information, which is called the \textit{optimally interacting theory}. Motivated by Bahrami's theory, we present a multi-classifier interactive learning (MCIL) method to address ambiguous data. Our goal is to identify an alternative that automatically constructs ambiguous labels (i.e. emotion probability distribution) instead of precise labels to ambiguous emotion data. In MCIL, multiple classifiers are firstly trained on a portion of ambiguous data using precise labels. Subsequently, they are used to vote for the other ambiguous data. Thus, the statistics of voting are used to construct ambiguous labels for retraining. Finally, these classifiers are retrained with ambiguous labels to interact and learn better information of ambiguous data, as shown in Fig.\ref{fig:tsnevisualization}.b, which mimics human interaction strategy.

Unlike previous methods, MCIL integrates varied information learned by different classifiers. It does not only improve the performance of each classifier, but also results in more consistent recognition results among multiple classifiers. Moreover, the classifiers in MCIL are firstly trained by the precise labels given by the database, then construct the ambiguous labels. This strategy enhances the generalization ability of MCIL, and make it also work on the databases without the information of experts’ voting.

Extensive experiments demonstrate that our proposed MCIL outperforms the state-of-the-art methods on three benchmarks, i.e. 3.1\% on MAS, 2.0\% on IEMOCAP, and 1.58\% on FAU-AIBO.
The main contributions of this work are following:

\begin{itemize}
	\item We propose multi-classifier interactive learning (MCIL), which uses multiple different classifiers to mimic human interacting behavior to address the ambiguous speech emotion recognition. MCIL can integrate varied information learned by different classifiers.
	\item MCIL firstly trains each classifier using precise labels given by the database, and then constructs the ambiguous labels. Thus, MCIL has a better generalization ability.
	\item Experimental results demonstrate that our proposed method raises both the performance and consistency among multiple classifiers.
\end{itemize}


\begin{figure*}[htb]	
	\begin{center}
		\includegraphics[width=16cm]{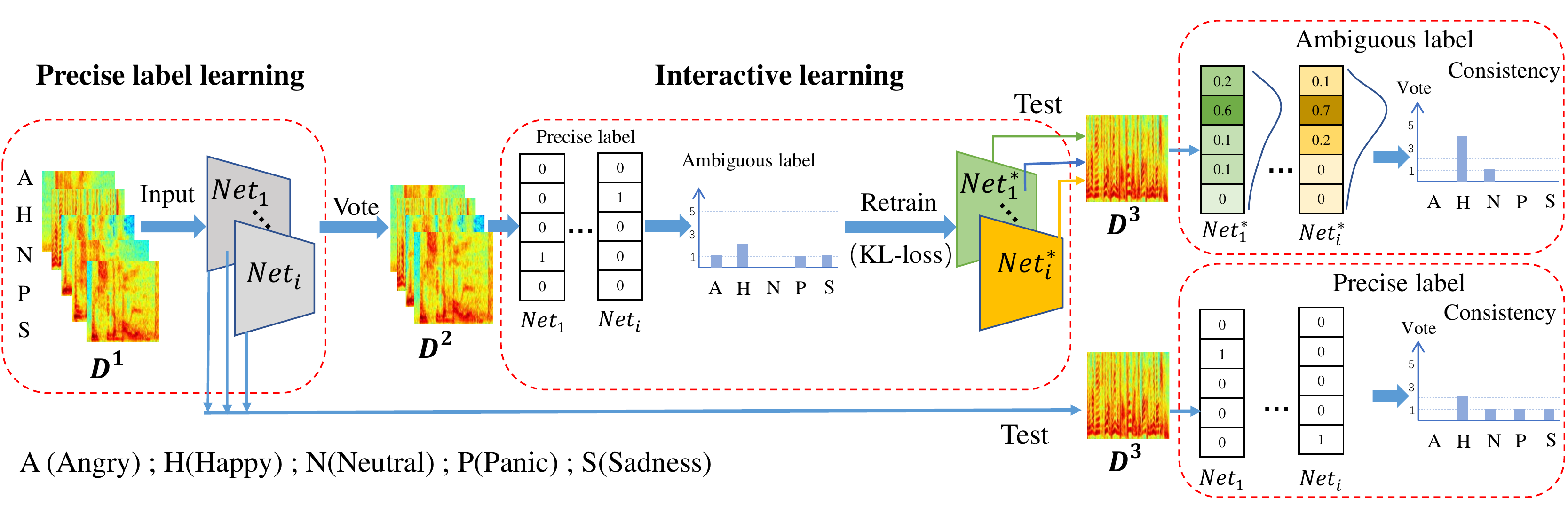}
	\end{center}
	\vspace{-.2in}
	\caption{An overview of our proposed MCIL framework for SER. In precise label learning, multiple classifiers simulate inconsistent cognition on precisely labeled $D^1$. And then they are retrained by the new ambiguous labels on $D^2$. After both stages, the models are tested on the dataset $D^3$. The output of interactive learning is a probability distribution, which more accurately represents ambiguous emotion data, and is more consistent among these classifiers.}
	\label{fig:structure}
\end{figure*}

\section{Related Work}
\label{se:relatedwork}

\subsection{Deep Learning Approachs in SER}

Since the first publication of the successful use of a convolutional neural network (CNN) for learning feature representations from speech signals \cite{mao2014learning}, several researchers have followed this trend to use deep neural networks to automatically learn feature representation \cite{gu2016speech}. Cummins \textit{et al.} \cite{cummins2017image} proposed a CNNs based method, that used a pre-trained AlexNet to extract deep spectrum features and used an SVM for classification. Li \textit{et al.} \cite{li2018attention} used two different convolution cores to extract temporal domain features and frequency domain features, then two different kinds of features were concatenated and fed to convolutional layers, and attention pooling was used in the last layer to increase accuracy. Wu \textit{et al.} \cite{wu2019speech} proposed to extract features with CNN, then combined a capsule network and a gated recurrent unit to address the classification task. Dai \textit{et al.} \cite{dai2019learning} combined Cross-Entropy loss and center loss to enhance the discriminating power of their proposed approach. All these methods use precise labels as the ground truth.

Unlike the above methods, a few recent studies suggested that the precise labels could not well represent the ambiguity of speech emotions, and then attempted to address this issue. Lotfian \textit{et al.} \cite{lotfian2018predicting} proposed a multi-task learning framework by specifying secondary emotions in addition to the dominant emotion, but was lack of other minor emotions. To solve this problem, soft-target label was presented by Ando \textit{et al.} \cite{ando2018soft}, which described the reference intensities of the target category. However, it assumed that all emotions existed in an utterance, and then estimated their proportions. This strategy made soft-target learning and computation complicated. To simplify the problem, Ando \textit{et al.} \cite{ando2019speech} proposed multi-label learning to represent all emotions that existed in utterance, instead of the emotion distribution. We can see all the above methods cannot represent the ambiguity well. Moreover, these methods heavily depend on the statistics of experts’ voting, thus, limits their generalization ability. On the contrast, MCIL can be applied on all types of datasets.

\subsection{Multi-Classifier Approaches}

To achieve a robust performance on complicated problems, many studies tried to mimic the collective decision-making behavior of human, such as ensemble learning \cite{dietterich2000ensemble}, co-training \cite{blum1998combining}, and tri-training \cite{zhou2005tri}.

Ensemble learning was designed to combine the decisions from diverse classifiers to improve the overall performance. The underlying idea is that even if one classifier gets a wrong prediction, other classifiers are still able to correct the error to maintain the performance (i.e. bagging \cite{breiman1996bagging}, boosting \cite{schapire1990strength} and stacking \cite{wolpert1992stacked}). However, there is no interaction between classifiers, which indicates that classifiers cannot improve their individual performance through the ensemble learning process.

Instead, co-training and tri-training, two typical semi-supervised algorithms, improve the performance of each classifier by interacting data among classifiers. In co-training algorithm, firstly, two classifiers are trained on a labeled dataset with two different views. Afterwards, two classifiers swap the samples according to their high confident predictions during semi-supervised procedure, in which these samples are treated as the new training data. Inspired by co-training, tri-training was proposed to train three classifiers on three training subsets, which are obtained by bootstrap sampling from labeled datasets. In semi-supervised procedure, each new sample is predicted by two classifiers. If the predicted labels are identical, this sample will be marked as the training data for the third classifier with this label. Due to above mechanism, one can see that only the discriminating data are involved in training and interacted, but those ambiguous data usually are excluded in the training process.

In contrast, MCIL improves the performance of classifiers by interacting their decisions, specifically, constructing the ambiguous label distribution as the new data label. It brings two merits: 1) a better representation of ambiguous data; 2) all data are included in the training procedure.

\section{Mathematically Analyzing the effectiveness of Optimal Interactive Theory}

The optimal interactive theory \cite{bahrami2010optimally} says that multiple individuals can come to an optimal joint decision by sharing information with each other. It has been proved by several psychological experiments that individuals with similar cognition abilities can perform better in an interactive process. Here, we try to mathematically analyze the effectiveness of the theory.

\begin{figure}[t]	
	\begin{center}
		\includegraphics[width=5.5cm]{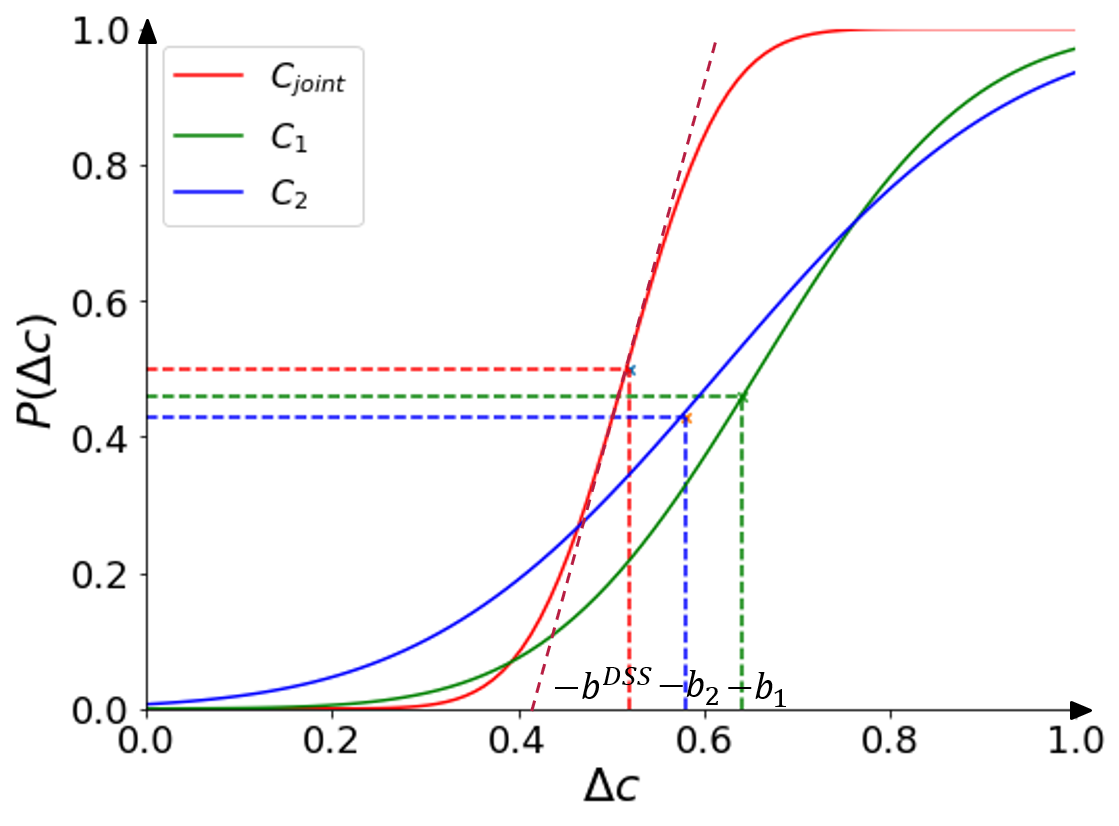}
	\end{center}
	\vspace{-.2in}
	\caption{The psychometric curves of participants (classifiers). The $x$ axis indicates the clarity of the data, $\Delta c$, and the $y$ axis indicates the cognition accuracy $P$ of the participants (classifiers). The $C_1$ and $C_2$ denote the curves of classifiers without interactive learning. The $C_{joint}$ is the joint decision curve after interactive learning, which shows a better performance on the ambiguous data.}
	\label{fig:learningcurve}
\end{figure}

In above experiments, psychologists constructed a psychometric curve (a cumulative Gaussian function\footnote{According to the central limit theorem, when the size of samples is large enough, the distribution of these samples tends to a Gaussian distribution. Therefore, many common phenomena can be represented by a Gaussian distribution or a cumulative Gaussian distribution.}) according to each participant that plots the cognition performance, $P$, of each individual against the clarity of the data, $\Delta c$, in the task, as shown in Fig. \ref{fig:learningcurve}. The curve is also determined by the participant's cognition ability, $\sigma$, and data clarity bias, $b$. The $\sigma$ is the variance of the psychometric curve, hence, denotes the participant's ability of making the right decision. The smaller $\sigma$ is, a better cognition performance achieves. The $b$ denotes the clarity of the data when the individual's cognition accuracy rises the fastest. A smaller $b$ indicates that the individual is more sensitive to the data with a lower clarity, thus, can make a better decision. Analogously, in machine learning, the performance of a classifier on ambiguous data could be also represented as a cumulative Gaussian function. The accuracy of each classifier, $P$, is determined by the classification capability of the classifier, $\sigma$, the clarity of the data, $\Delta c$, and the data clarity bias $b$,

\begin{equation}
	P(\Delta c)= H(\frac{{\Delta c}+b}{\sigma}),
\end{equation}
where,

\begin{equation}
	H\left( z \right)=\int_{-\infty }^{z}{\frac{dt}{{{\left( 2\pi  \right)}^{{1}/{2}\;}}}}\exp \left[ {-{{t}^{2}}}/{2}\; \right].
\end{equation}

Given two different classifiers $C_1$ and $C_2$, Figure ~\ref{fig:learningcurve} shows that their classification accuracies rise with the increase of clarity of the data, $\Delta c$. However, due to different classification capabilities, their accuracies increase with different rates. It should be noted that the maximum slope $S$ of the curve indicates the sensitivity of the classifier to the change of clarity of data. A larger $S$ indicates a smaller variance, which means a better classification performance. $S$ can be obtained by taking partial derivative of $\Delta c$ over $P(\Delta c)$, which is inversely proportional to the variance of curve, $\sigma$,

\begin{equation}
	S=\frac{1}{\sqrt{2\pi{\sigma^2}}}.
\end{equation}

Additional psychological studies \cite{ernst2002humans,alais2004ventriloquist} have shown that multiple participants will lead to a new psychometric curve after interactive learning, as the red curve shown in Fig. \ref{fig:learningcurve}. To prove it, we define a joint performance, $P^{joint}$, after interactive learning among multiple classifiers according to the psychological model in \cite{jogan2015signal},
\begin{equation}
	P^{joint}(\Delta c)= H(\frac{\Delta c + b^{joint}}{\sigma^{joint}}),
\end{equation}
where,

\begin{equation}
	b^{joint} = \sum_{i=1}^n w_i b_i, \qquad \sigma^{2^{joint}} = \sum_{i=1}^n w_i^2 \sigma_i^2,
	\label{eq:sigma-DSS}
\end{equation}
where, the weight is defined as
\begin{equation}
	w_i = \frac {\prod_{j=1}^n \sigma_j^2}{\sigma_i^2 \sum_{j=1}^n (\frac{\prod_{k=1}^n \sigma_k^2}{\sigma_j^2})}.
	\label{eq:weight}
\end{equation}

Taken the partial derivative of $\Delta c$ over $P^{joint}$ ($\Delta c$), $S^{joint}$ can be approximated by

\begin{equation}
	S^{joint} \approx \sqrt{\sum_{i=1}^n S_i^2}.
	\label{eq:S-DSS}
\end{equation}

According to the inequality relation, we have

\begin{equation}
	S^{joint} \geq max(S_i), \qquad i=1,\cdots,n.
\end{equation}

This result is consistent to the red curve in Fig. \ref{fig:learningcurve}. It can be concluded that the sensitivity, $S^{joint}$, of the classifier has been enhanced by interactive learning.

Moreover, according to Eq. \ref{eq:sigma-DSS} and Eq. \ref{eq:weight}, the variance $\sigma^{2^{joint}}$ after interactive learning can be written as

\begin{equation}
\sigma^{2^{joint}} = \frac {\prod_{i=1}^n \sigma_i^2}{\sum_{i=1}^n (\frac{\prod_{j=1}^n \sigma_j^2}{\sigma_i^2})}
\end{equation}

To evaluate the $\sigma^{2^{joint}}$, we subtract arbitrary $k$-th $\sigma_k^{2}$ that does not take interactive learning.

%

\begin{equation}
\begin{aligned}
\sigma^{2^{joint}} - \sigma_k^{2} &= \frac {\prod_{i=1}^n \sigma_i^2}{\sum_{i=1}^n (\frac{\prod_{j=1}^n \sigma_j^2}{\sigma_i^2})}  - \sigma_k^{2} \\
&= \frac {{\prod_{i=1}^n \sigma_i^2} - \sigma_k^{2}{\sum_{i=1}^n (\frac{\prod_{j=1}^n \sigma_j^2}{\sigma_i^2})}}{\sum_{i=1}^n (\frac{\prod_{j=1}^n \sigma_j^2}{\sigma_i^2})} \\
&\le  0 ,
\end{aligned}
\end{equation}
then, we have

\begin{equation}
\sigma^{2^{joint}} \leq min(\sigma_k^2), \qquad k=1,\cdots,n.
\end{equation}

This result also proves the classification capability of classifiers can be improved through interactive learning on the task of recognizing the ambiguous data.

The above analysis shows the effectiveness of interactive learning. It indicates that classifiers could simulate the interaction process between human beings. Therefore, the question becomes how to implement interactive learning for multiple classifiers. We then propose the MCIL algorithm.

\section{The Proposed Method}

In this section, we first describe the problem formulation of emotion ambiguity. Subsequently, the idea of constructing ambiguous labels is explained. Finally, the function of interactive learning is detailed. Figure~\ref{fig:structure} shows our framework.

\subsection{Problem Formulation}

In past studies, the precise labels were commonly used in SER task, which were mostly defined as $Y\in\{0,1,2,3,4\}$ to denote \textit{angry}, \textit{happy}, \textit{neutral}, \textit{panic}, and \textit{sadness}. However, far from the optimal situation, there are always some ambiguous data between emotions in the real world. Therefore, it is not always feasible to use precise labels to represent these data.

To solve this issue, we begin designing ambiguous labels for emotion, which are defined as follows:

\begin{equation}
	\widetilde Y_x=(y_x^a, y_x^h, y_x^n, y_x^p, y_x^s|y_x^a+y_x^h+y_x^n+y_x^p+y_x^s=1),
\end{equation}

\noindent where, $y_x^a, y_x^h, y_x^n, y_x^p, y_x^s \in[0,1]$ represent the probability that the ambiguous degree of emotion $x$ belongs to \textit{angry}, \textit{happy}, \textit{neutral}, \textit{panic}, and \textit{sadness} category, respectively. Therefore, $\widetilde Y_x$ indicates that a sample is suggested to be in a category with a higher probability label.

In practice, ambiguous distribution ($y_x^a, y_x^h, y_x^n, y_x^p, y_x^s$) is usually unknown and difficult to obtain. Ideally, we can invite $N$ participants to manually annotate a precise label for these emotional data. Thus, an ambiguous label can be obtained by their voting:

\begin{equation}
	\label{eq:ambiguouslabel}
	(\frac{v_x^a}{N},\frac{v_x^h}{N},\frac{v_x^n}{N},\frac{v_x^p}{N},\frac{v_x^s}{N}),
\end{equation}

\noindent where $v$ represents the number of votes for each category on the sample $x$. However, this is time-consuming and requires a large workload. Therefore, we present a novel idea of using multiple classifiers to simulate different individual's inconsistent cognition to vote for these samples.

\subsection{The Construction of Ambiguous Labels}

In optimally interacting theory, individuals are able to get a better solution to a complicated problem by communicating with others. To mimic this process, we employ $N$ neural networks \{$Net_i\}$ with different architectures for training and testing on emotional speech datasets to construct ambiguous labels for interactive communication.

To this end, we design three datasets $D^1$, $D^2$, and $D^3$, respectively. $D^1$ contains the samples with precise labels, which are less ambiguous and used to train $N$ classifiers in the precise label learning stage. $D^3$ consists of ambiguous samples and is employed to evaluate the performance of interactive learning. $D^2$ includes a large number of samples that are unlabeled, and is applied to ambiguous label construction and subsequent retraining.

First, as shown in the left section of Fig.\ref{fig:structure}, all classifiers are trained on $D^1$ for precise label learning. Here, we use the Cross-Entropy to optimize the $\{Net_i\}$, which is defined as follows,

\begin{equation}
	Loss_{CE}=\sum_{j=1}^m\sum_{i=1}^n -y_{ji}log(\hat y_{ji})-(1-y_{ji})log(1-\hat y_{ji}),
\end{equation}

\noindent where $n$ denotes the number of emotions, $m$ denotes the number of the samples, $y_{ji}$ denotes the ground truth, and $\hat y_{ji}$ represents the output of $Net_i$.

Since each $Net_i$ has a different architecture, the learned representation is different from that of the others. Furthermore, each category of ambiguous emotion brings uncertain knowledge to each $Net_i$. Thus inconsistent cognition among human beings can be mimicked.




Consequently, the well-trained $\{Net_i\}$ are used to classify the unlabeled data in $D^2$. Then, the classification results of $N$ networks obtained from each sample, are counted by voting and normalized to be ambiguous labels.


The label of a sample $x$ is denoted as $\widetilde Y_x = (y^a_x, y^h_x, y^n_x, y^p_x, y^s_x)$, which represents the probability that $x$ belongs to each category. As the results from the different knowledges of $N$ networks, no manual annotation is required. Meanwhile, we regard the constructed labels as the results of communication among these networks.


\subsection{Interactive Learning}


At this moment, the data in $D_2$ have their new ambiguous labels $\widetilde Y_x$, which contain slightly different information from precise labels. Afterwards, we are able to retrain $\{Net_i^*\}$ on $D_2$ to minimize the gap between the predicted label and the label $\widetilde Y_x$ given by $\{Net_i\}$, as shown in the middle section of Fig.\ref{fig:structure}. We call this \textit{interactive learning}.

Since ambiguous labels are a probability distribution, we select KL divergence to represent them. Therefore, $\{Net_i^*\}$ are optimized by KL loss during the retraining stage, which is defined below,


\begin{equation}
	\label{eq:KLdivergence}
	Loss_{KL}=\sum_j^m \widetilde Y_j\log\frac{\widetilde Y_j}{\hat{Y}_j},
\end{equation}

\noindent where $\hat Y$ denotes the predicted label, $\widetilde Y$ denotes the ambiguous label given by $\{Net_i\}$, and $m$ denotes the number of samples.

This procedure mimics how human beings learn new knowledge by communicating with others. Similarly, each $\{Net_i^*\}$ also learns more comprehensive information of data from other networks. Other studies use precise labels for recognition, in other words, each sample belongs to only one emotional category in their methods. In contrast with them, our output is a distribution that may span more than one category.  In order to be consistent with other studies, we choose the highest probability in the distribution as final output category as illustrated in the right section of Fig.\ref{fig:structure}.

Since the conclusion of Bahrami's paper \cite{bahrami2010optimally} pointed that ``The general consensus from extensive earlier work on collective decision-making is that groups rarely outperform their best members", we follow the consensus and choose the best classifier of MCIL to make the final decision.


\section{Experimental Setting}

\subsection{Database}

To evaluate MCIL, we conduct the performance test on three benchmark databases.

\noindent \textbf{Mandarin Affective Speech (MAS)}. Since Chinese culture and emotional expression are conservative, which cause Chinese pronunciation and intonation to be mild, the features of data in Chinese corpora are somewhat ambiguous. The speakers expressed 5 different emotions including \textit{angry}, \textit{happy}, \textit{neutral}, \textit{panic} and \textit{sadness} to act the utterance. Speakers read the same sentence three times repeatedly and 20 sentences in total for each emotion.

\noindent \textbf{Interactive Emotional Dyadic Motion Capture (IEMOCAP)}. IEMOCAP corpus \cite{busso2008iemocap} was a commonly used English corpus for SER. Following \cite{luo2018investigation,mao2019deep}, we used both improvised and scripted data, and chose \textit{angry}, \textit{happy}, \textit{neutral}, \textit{sadness}  and \textit{excited} as the basic emotions. With reference to \cite{busso2008iemocap,dai2019learning}, we merged \textit{happy} and \textit{excited} as \textit{happy} since they are close in the activation and valence domain.

\noindent \textbf{FAU-AIBO Emotion Corpus (FAU-AIBO)}. FAU-AIBO is a corpus of German children communicating with Sony’s AIBO pet robot \cite{batliner2008releasing}. The corpus can be divided into 2 or 5 emotional categories. To verify the robustness of MCIL, we chose the 2-emotion, which are NEG(active): states with negative valence (angry, torchy, reprimand, emphatic) and IDL(e): all other states. The details of three corpora are summarized in Table~\ref{tab:emotiondatasummary}.

\begin{table} [tbh]
	\caption{A brief overview of the MAS, IEMOCAP and FAU-AIBO corpora.}
	\label{tab:emotiondatasummary} \centering
	\vspace{1mm}
	\setlength{\tabcolsep}{1mm}{
		\centerline{
			\begin{tabular}{lccccc}
				\hline
				Corpora &  Language &  Utterance  & Subjects  & Emotion   \\ \hline
				MAS & Chinese  & 20400  &   68 (23 female) &  5      \\
				IEMOCAP & English  & 5531  &   10 (5 female) &  4    \\
				FAU-AIBO & German & 18216 & 51(30 female) & 2 \\			
				\hline
	\end{tabular}}}
\end{table}


\subsection{Evaluation Metrics}

The following two metrics are employed to evaluate the effectiveness of MCIL on speech emotion recognition:

\noindent \textbf{Classification accuracy}. We evaluated the performance of each classifier with respect to recognition accuracy tested on $D^3$.

\noindent \textbf{Consistency among multiple classifiers}. Due to the different classifier architectures and ambiguous data, the recognition results of $\{Net_i\}$ and $\{Net_i^*\}$ should not be consistent. One of the purposes of interactive learning is to improve classification consistency among all classifiers. Thus, a Kappa value was used to evaluate the consistency of classifiers:

\begin{equation}
	K=\frac{\overline{P}-\overline{P}_e}{1-\overline{P}_e},
\end{equation}
where,

\begin{equation}
	\overline{P}=\frac{1}{d}\sum_{i=1}^d P_i, \quad \overline{P}_e=\sum_{j=1}^c P^2_j,
\end{equation}

\begin{equation}
	P_i=\frac{1}{n(n-1)}\sum_{j=1}^c (v_{ij}^2 - n), \quad P_j=\frac{\sum_{i=1}^d v_{ij}}{d*n},
\end{equation}

\begin{table} [b!]
	\caption{The sizes of training and testing datasets.}
	\label{trainandtestdataset} \centering
	\setlength{\tabcolsep}{6mm}{
		\centerline{
			\begin{tabular} {l c c c}
				\hline
				Corpora & $D^1$ & $D^2$ & $D^3$ \\
				\hline
				MAS   & 6000 & 13400 &  1000   \\
				IEMOCAP & 1710 & 3421 & 400\\
				FAU-AIBO & 5375 & 10750 & 2091   \\
				\hline
	\end{tabular}}}
\end{table}

\begin{table*}[t!]
	\caption{The accuracies of five classifiers with and without the interactive learning on MAS (grey background) and IEMOCAP (white background).}
	\centering
	\small
	\resizebox{\textwidth}{26mm}{
		\begin{tabular}{c|cc|cc|cc|cc|cc|cc}
			\hline		
			&   \multicolumn{2}{c|}{Angry} & \multicolumn{2}{c|}{Happy}    & \multicolumn{2}{c|}{Neutral}  &  \multicolumn{2}{c|}{Panic}  &  \multicolumn{2}{c|}{Sadness} &  \multicolumn{2}{c}{Overall}  \\ \hline
			Methods & Baseline & MCIL & Baseline & MCIL & Baseline & MCIL & Baseline & MCIL & Baseline & MCIL & Baseline & MCIL \\
			\multirow{2}*{DenseNet} & \cellcolor{mygray}{57.0\%} &	\cellcolor{mygray}{\textbf{61.5}\%} &	\cellcolor{mygray}{48.5\%} &	\cellcolor{mygray}{\textbf{52.5}\%} &	\cellcolor{mygray}{73.0\%} &	\cellcolor{mygray}{\textbf{78.5}\%} &\cellcolor{mygray}{57.5\%} &	\cellcolor{mygray}{\textbf{59.5}\%} &	\cellcolor{mygray}{\textbf{66.0}\%} &	\cellcolor{mygray}{59.0\%} &	\cellcolor{mygray}{60.4\%} & \cellcolor{mygray}{\textbf{62.2}\%}  \\				
			~ & \textbf{71.0}\% & 64.0\% &49.0\% & \textbf{52.0}\% &73.0\% & \textbf{74.0}\% & \multicolumn{1}{c}{\rule[0.05cm]{0.3cm}{0.03mm}} & \multicolumn{1}{c|}{\rule[0.05cm]{0.3cm}{0.03mm}} & 61.0\% & \textbf{78.0}\% & 63.5\% & \textbf{67.0}\% \\
			\multirow{2}*{ResNet} & \cellcolor{mygray}{58.0\%} &	\cellcolor{mygray}{\textbf{61.5}\%} &	\cellcolor{mygray}{43.5\%} &	\cellcolor{mygray}{\textbf{47.0}\%} &	\cellcolor{mygray}{68.5\%} &	\cellcolor{mygray}{\textbf{79.0}\%} &	\cellcolor{mygray}{55.5\%} &	\cellcolor{mygray}{\textbf{62.5}\%} &	\cellcolor{mygray}{\textbf{59.0}\%} &	\cellcolor{mygray}{57.0\%} &	\cellcolor{mygray}{56.9\%} &	\cellcolor{mygray}{\textbf{61.4}\%}   \\
			~ & 64.0\% & \textbf{69.0}\% & \textbf{51.0}\% & 50.0\% & 60.0\% & \textbf{71.0}\% &\multicolumn{1}{c}{\rule[0.05cm]{0.3cm}{0.03mm}} &\multicolumn{1}{c|}{\rule[0.05cm]{0.3cm}{0.03mm}}  & 67.0\% & \textbf{79.0}\% & 60.5\% & \textbf{67.3}\% \\[3pt]
			\multirow{2}*{VGG} & \cellcolor{mygray}{53.5\%} &	\cellcolor{mygray}{\textbf{59.0}\%} &	\cellcolor{mygray}{41.5\%} &	\cellcolor{mygray}{\textbf{46.5}\%} &	\cellcolor{mygray}{70.5\%} &	\cellcolor{mygray}{\textbf{77.5}\%} &	\cellcolor{mygray}{48.5\%} &	\cellcolor{mygray}{\textbf{57.0}\%} &	\cellcolor{mygray}{58.5\%} &	\cellcolor{mygray}{\textbf{61.5}\%} &	\cellcolor{mygray}{54.5\%} &	\cellcolor{mygray}{\textbf{60.3}\%}    \\
			~ & 63.0\%& 63.0\% & \textbf{59.0}\% & 50.0\%  & 62.0\% & \textbf{73.0}\% & \multicolumn{1}{c}{\rule[0.05cm]{0.3cm}{0.03mm}} &\multicolumn{1}{c|}{\rule[0.05cm]{0.3cm}{0.03mm}} & 69.0\% & \textbf{78.0}\% & 63.0\% & \textbf{66.0}\% \\
			\multirow{2}*{AlexNet} & \cellcolor{mygray}{54.0\%} &	\cellcolor{mygray}{\textbf{59.0}\%} &	\cellcolor{mygray}{43.0\%} &	\cellcolor{mygray}{43.0\%} &	\cellcolor{mygray}{70.5\%} &	\cellcolor{mygray}{\textbf{77.0}\%} &	\cellcolor{mygray}{52.5\%} &	\cellcolor{mygray}{\textbf{60.5}\%} &	\cellcolor{mygray}{\textbf{63.0\%}} &	\cellcolor{mygray}{53.5\%} &	\cellcolor{mygray}{56.6\%} &	\cellcolor{mygray}{\textbf{58.6}\%}    \\
			~ & 55.0\% & \textbf{58.0}\% & \textbf{57.0}\% & 55.0\% &58.0\% & \textbf{66.0}\% & \multicolumn{1}{c}{\rule[0.05cm]{0.3cm}{0.03mm}} &\multicolumn{1}{c|}{\rule[0.05cm]{0.3cm}{0.03mm}} & 64.0\% & \textbf{77.0}\% &58.5\% & \textbf{64.0}\% \\[3pt]
			\multirow{2}*{SqueezeNet} & \cellcolor{mygray}{55.0\%} &	\cellcolor{mygray}{\textbf{59.5}\%} &	\cellcolor{mygray}{46.5\%} &	\cellcolor{mygray}{43.5\%} &	\cellcolor{mygray}{65.5\%} &	\cellcolor{mygray}{\textbf{74.0}\%} &	\cellcolor{mygray}{56.5\%} &	\cellcolor{mygray}{\textbf{64.5}\%} &	\cellcolor{mygray}{58.0\%} &	\cellcolor{mygray}{\textbf{59.0}\%} &	\cellcolor{mygray}{56.3\%} &	\cellcolor{mygray}{\textbf{60.1}\%}   \\
			~ & 63.0\% & 63.0\% &  \textbf{59.0}\%& 57.0\% & 68.0\%  & \textbf{69.0}\% &  \multicolumn{1}{c}{\rule[0.05cm]{0.3cm}{0.03mm}} &\multicolumn{1}{c|}{\rule[0.05cm]{0.3cm}{0.03mm}}  & 66.0\% & \textbf{81.0}\% & 64.0\% & \textbf{67.0}\% \\
			\hline
	\end{tabular}}
	\label{tab:semisupervisedlearning}
\end{table*}
\noindent where, $d$ denotes the number of test samples, $n$ denotes the number of classifiers, $c$ denotes the number of categories, $\{v_{ij}|i=1,\cdots,d; j=1,\cdots,c\}$ denotes the number of votes for the sample $i$ is classified as category $j$, $P_i$ denotes the consistency for each sample $i$, and $P_j$ denotes the consistency for each category.

\subsection{Implementation Details}

The five classifiers in the proposed framework were the 34-layer ResNet \cite{he2016deep}, 121-layer DenseNet \cite{huang2017densely}, SqueezeNet \cite{iandola2016squeezenet}, 11-layer VGG \cite{simonyan2014very}, and AlexNet \cite{krizhevsky2012imagenet}. For DenseNet, ResNet, AlexNet, and VGG, according to the number of categories in FAU-AIBO, IEMOCAP, and MAS, the last fully-connected layer was adjusted to two, four, and five outputs, respectively. For SqueezeNet, the output dimension of the final convolution layer was reduced to two, four, and five, respectively. For each classifier, we adjusted the number of layers to avoid over-fitting.

The training procedure of all classifiers was implemented using PyTorch on an NVIDIA 2080Ti GPU, and all five $Net_i$ were pre-trained. The corpora were divided into three groups as shown in Table \ref{trainandtestdataset}. $D^1$, which contained samples with precise labels that are less ambiguous than those in the other groups, was used to train $Net_i$ in the precise label learning stage. To avoid over-fitting, we expand $D^1$ by horizontally flipping the samples as data augmentation. $D^2$, which contained unlabeled samples, was used to construct the ambiguous label and retrain the five $Net_i^*$. Finally, $D^3$, which contained labeled but more ambiguous samples, was used to evaluate the performance of classifiers.

\begin{table}[b!]
	\vspace{-.1in}
	\caption{The accuracies of five classifiers with and without interactive learning on FAU-AIBO.}
	\label{tab:FAU-AIBO} \centering

	\centerline{
		\resizebox{88mm}{12.5mm}{
			\begin{tabular} {l|cc|cc|cc}
				\hline			
				& \multicolumn{2}{c|}{IDL} & \multicolumn{2}{c|}{NEG} & \multicolumn{2}{c}{Overall}  \\
				\hline
				Methods & Baseline & MCIL  & Baseline & MCIL  & Baseline & MCIL  \\
				DenseNet & 76.11\% & \textbf{77.48\%} & 46.24\% & \textbf{48.00\%} & 63.94\% & \textbf{65.47\%} \\
				ResNet & 74.25\% & \textbf{77.08\%} & 45.54\% & \textbf{48.36\%} & 62.55\% & \textbf{65.38\%}  \\
				VGG   & 74.41\% & \textbf{75.87\%} & 48.94\% & \textbf{49.53\%} & 64.04\% & \textbf{65.14\%} \\
				AlexNet & 76.51\% & \textbf{76.59\%} & 45.49\% & \textbf{46.24\%} & 63.86\% & \textbf{64.23\%}  \\
				SqueezeNet & 60.45\% & \textbf{68.85\%} & 39.79\% & \textbf{53.76\%} & 52.03\% & \textbf{62.70\%}  \\	
				\hline
	\end{tabular}}}
\end{table}

\noindent \textbf{Precise label learning:} To train these five $Net_i$ for precise labels, we applied 5-fold cross-validation on $D^1$. Since random cropping or resizing may influence the recognition of spectrograms, all samples were used with the original size in this work. During training, the parameters of the first few layers were fixed, and the subsequent layers were optimized. The objective function was Cross-Entropy loss. These models were trained using Adam with a batch size of 16, and the learning rate decaying exponentially from $10^{-4}$ to $10^{-8}$. After training, five $Net_i$ were tested on $D^3$, and their results were used as the baseline.

\noindent \textbf{Interactive learning:} First, five trained $Net_i$ classified the unlabeled samples in $D^2$. Their results constructed the ambiguous label $\widetilde Y_x$ for each sample $x$. Then, these $Net_i^*$ were retrained on $D^2$ using the new label $\widetilde Y$. The classifiers were optimized by minimizing the new objective function KL-loss. We fixed the first few layers of each classifier and retrained them with a batch size of 16, weight decay of 0.0005, and learning rate decaying exponentially from $10^{-4}$ to $10^{-8}$ across 50 epochs. The retraining was stopped when no change of KL-loss value.

\section{Experimental Results Analysis}

\subsection{Ablation studies: the efficiency of 3, 5, and 7 classifiers in MCIL}
To analyze the influence of the number of the classifiers used in MCIL, we further evaluate the using of 3, 5, and 7 classifiers on IEMOCAP that is widely used. The results are listed in table~\ref{tab:IEMOCAP}.

As shown in the table, using 5 classifiers to train MCIL achieves the best results. When we increase the number of classifiers from 3 to 5, we can observe clear improvements from 2.75\% to 6.75\% for all classifiers. Moreover, four of five classifiers (i.e ResNet, VGG, SqueezeNet, and DenseNet) obtain the best performance. The reason behind could be that 5 classifiers bring more diversities of the classifiers with inconsistent cognition than 3 classifiers. This leads to a better representation of ambiguous data, and further improve the consistency and accuracy of all classifiers. However, this does not mean increasing the number of classifier will always get a better results. As we can see, 7-classifer does not outperform 5-classifier. The explanation might be that the two additional have lower performances than that of other classifiers in baseline. They weaken the cognitive ability of the entire group of all classifiers, thus, lead to the decreasement of the overall accuracy.

\begin{table}[htbp]
	\centering                                                                      	\caption{The performance of 3, 5, and 7 classifiers on IEMOCAP.}                                                                      	\resizebox{88mm}{17mm}{
		\begin{tabular}{lcccc}
			\toprule
			& \multicolumn{1}{c}{Baseline} & \multicolumn{1}{c}{3-classifier} & \multicolumn{1}{c}{5-classifier} & \multicolumn{1}{c}{7-classifier} \\
			\midrule
			ReseNet & 60.50\% & 67.00\% & \textbf{67.25\%} &\textbf{ 67.25\%} \\
			VGG   & 63.25\% & 65.20\% & \textbf{66.00\%} & 65.00\% \\
			SqueezeNet & 64.00\% & 65.75\% & \textbf{67.00\%} & 64.25\% \\
			DenseNet & 63.50\% & \multicolumn{1}{c}{\rule[0.05cm]{0.3cm}{0.03mm}}  & \textbf{67.00\%} & 65.00\% \\
			AlexNet & 58.50\% & \multicolumn{1}{c}{\rule[0.05cm]{0.3cm}{0.03mm}} & 64.00\% & \textbf{64.25\%} \\
			MobileNet & 61.50\% & \multicolumn{1}{c}{\rule[0.05cm]{0.3cm}{0.03mm}} & \multicolumn{1}{c}{\rule[0.05cm]{0.3cm}{0.03mm}} & 63.75\% \\
			Inception & 60.20\% & \multicolumn{1}{c}{\rule[0.05cm]{0.3cm}{0.03mm}} &  \multicolumn{1}{c}{\rule[0.05cm]{0.3cm}{0.03mm}} & 60.75\% \\
			\bottomrule
	\end{tabular}}%
	\label{tab:IEMOCAP}%
\end{table}%

\subsection{The effectiveness of the MCIL method}

To evaluate the performance of our proposal, we first test the five classifiers for ambiguous emotion recognition with and without interactive learning. Table~\ref{tab:semisupervisedlearning} lists their performance changes on $D^3$ of MAS and IEMOCAP.

As we can see, on both corpora, the overall accuracy of five $Net_i^*$ are entirely improved. Firstly, we analyze the performance on the most ambiguous emotion categories \textit{angry} and \textit{happy} in Mandarin, which obtain the lower performance among five emotions. Although, VGG obtains 53.5\% and 41.5\% accuracies of \textit{angry} and \textit{happy} when trained with the precise label, respectively, MCIL introduces the remarkable improvements of 5.5\% and 5\%. The five $Net_i^*$ all lead to performance improvement on ambiguous emotion \textit{angry}. Secondly, we see that the \textit{neutral} emotion achieves the highest accuracy among all emotion categories. ResNet and SqueezeNet can obtain 10.5\% and 8.5\% improvement on the \textit{neutral} emotion, respectively. The overall accuracy of VGG rises remarkably by 5.8\%, and DenseNet increases from 60.4\% to 62.2\% as well.

Meanwhile, the results from the IEMOCAP corpus are similar to the MAS. The overall accuracy of ResNet achieves a remarkable 6.8\% increase, and AlexNet rises from 58.5\% to 64.0\%. DenseNet, VGG, and SqueezeNet obtain 3.5\%, 3.0\%, and 3.0\% improvements, respectively.

Table~\ref{tab:FAU-AIBO} lists the performance of FAU-AIBO. Since FAU-AIBO is severely unbalanced, specifically, IDL is much bigger than NEG in terms of both total sample size and clear sample size, all classifiers perform better on IDL. After interactive learning, we can observe all five networks are improved on both categories. Particularly, the accuracy of SqueezeNet rises 8.4\% on IDL, and rises 13.97\% on NEG, and finally results in a 10.63\% improvement. Other classifiers also achieve an improvement from 0.37\% to 10.66\%. For more details, please refer to supplementary material.

The above results suggest that, even though ambiguous labels are constructed by five classifiers' voting, they can better represent the ambiguous data. Interactive learning endows each ${Net_i^*}$ ability to transfer more useful and comprehensive information of ambiguous data to other networks.

\subsection{Confusion matrix}
\begin{table*}[hbtp]
	\centering
	\caption{Confusion matrix of DenseNet on MAS with and without interactive learning.}
	\begin{tabular}{l|cc|cc|cc|cc|cc}
		\hline
		& \multicolumn{2}{c|}{Angry} & \multicolumn{2}{c|}{Happy} & \multicolumn{2}{c|}{Neutral} & \multicolumn{2}{c|}{Panic} & \multicolumn{2}{c}{Sadness} \\
		\hline
		& Baseline & MCIL  & Baseline & MCIL  & Baseline & MCIL  & Baseline & MCIL  & Baseline & MCIL \\
		Angry & 57.00\% & \textbf{61.50\%} & 21.00\% & 21.00\% & 6.00\% & 4.50\% & 13.50\% & 11.50\% & 2.50\% & 1.50\% \\
		Happy & 17.00\% & 15.50\% & 48.50\% & \textbf{52.50\%} & 8.50\% & 10.00\% & 15.50\% & 17.50\% & 10.50\% & 4.50\% \\
		Neutral & 4.00\% & 5.50\% & 4.50\% & 4.00\% & 73.00\% & \textbf{78.50\%} & 1.50\% & 1.00\% & 17.00\% & 11.00\% \\
		Panic & 8.50\% & 7.50\% & 17.00\% & 20.00\% & 5.50\% & 6.00\% & 57.50\% & \textbf{59.50\%} & 11.50\% & 7.00\% \\
		Sadness & 2.00\% & 1.50\% & 0.50\% & 0.50\% & 22.50\% & 29.50\% & 9.00\% & 9.50\% & \textbf{66.00\%} & 59.00\% \\
		\hline
	\end{tabular}%
	\label{tab:confusionMAS}%
\end{table*}%

\begin{table*}[htbp]
	\caption{Confusion matrix of DenseNet on IEMOCAP with and without interactive learning.}
	\centering
	\small
	\begin{tabular}{l|cc|cc|cc|cc}		
		\hline
		\multicolumn{1}{l|}{Emotion} &   \multicolumn{2}{c|}{Angry} & \multicolumn{2}{c|}{Happy}    & \multicolumn{2}{c|}{Neutral}   &  \multicolumn{2}{c}{Sadness}   \\\hline
		& Baseline & MCIL  & Baseline & MCIL & Baseline & MCIL & Baseline & MCIL \\
		Angry & \textbf{71.00}\% &	64.00\% &	12.00\% &	17.00\% &	15.00\% &	16.00\% &	2.00\% &	3.00\%   \\
		Happy & 15.00\% &	12.00\% &	49.00\% &	\textbf{52.00}\% &	30.00\% &	32.00\% &	6.00\% &	4.00\%   \\
		Neutral  & 3.00\% &	1.00\% &	18.00\% &	14.00\% & 73.00\% &	\textbf{74.00}\% &6.00\% &	11.00\%   \\
		Sadness & 2.00\% &	2.00\% &	17.00\% &	9.00\% &	20.00\% &	11.00\%  &	61.00\% &	\textbf{78.00}\%   \\  \hline
	\end{tabular}
	\label{tab:confusion matrix IEMOCAP}
\end{table*}

We further evaluate the performance of MCIL on each emotion category in MAS and IEMOCAP. Table~\ref{tab:confusionMAS} and table~\ref{tab:confusion matrix IEMOCAP} list the confusion matrix of DenseNet with and without the interactive learning.


On MAS, four emotion categories have improvement after interactive learning. As we can see, on most ambiguous emotion categories \textit{angry} and \textit{happy} in Mandarin, there is a 1.5\% decrease of misclassifying from \textit{happy} to \textit{angry}. And the most unique emotion \textit{neutral} achieves the highest accuracy 78.5\% among all emotion categories. We can also observe a 5.5\% accuracy increase on \textit{neutral} when using interactive learning. Finally, \textit{sadness} is the only emotion category that decreases in accuracy. We have discovered that during the stage of ambiguous labels construction, some ambiguous labels that should belong to \textit{sadness} are more inclined to \textit{neutral}. Consequently, this causes a decrease in the accuracy of \textit{sadness}.

On IEMOCAP, the performances of three of the four emotion categories are improved. Specifically, 3.0\% and 1.0\% accuracy increases on \textit{happy} and \textit{neutral} can be observed after using interactive learning. Different from the MAS, the \textit{sadness} achieves a remarkably 17\% increase from 61.0\% to 78.0\%, which is also the highest accuracy 78.0\% among all emotion categories. The only exceptional emotion category that decreases 7.0\% in accuracy is \textit{happy}. We can also observe that there is a 5.0\% misclassification increase from \textit{angry} to \textit{happy}.
\subsection{Comparison of visualized feature representations}

To gain insight into MCIL, we extended the effectiveness examination at the feature representation level. Each emotion category in $D^3$ of MAS database was visualized by t-SNE in Fig.~\ref{fig:tsnevisualization} using the learned feature representations of VGG net with and without employing MCIL.

Figure~\ref{fig:tsnevisualization}.a denotes the feature representations trained with only precise label learning, treated as a baseline. Figure~\ref{fig:tsnevisualization}.b denotes the feature representations with interactive learning. There exist clearly less overlaps between categories in Fig. \ref{fig:tsnevisualization}.b compared with Fig. \ref{fig:tsnevisualization}.a. This illustrates that VGG net obtains a stronger separability on ambiguous emotions after employing MCIL. To have a quantitative comparison, we compute the normalized inner-class distances of feature representations with and without employing MCIL. The result shows the distance decreases from 0.663 of baseline to 0.638 of MCIL, which means that the feature representations of MCIL are more compact and consistent.

\subsection{Comparison with state-of-the-art}

\begin{table}[b]
	\centering
	\caption{The comparison with state-of-the-art methods}
	\small
	\resizebox{78mm}{20mm}{
		\begin{tabular}{lccc}
			\hline
			\multicolumn{4}{c}{Overall Accuracy} \\
			\cmidrule{2-4}
			& MAS & IEMOCAP & FAU-AIBO \\	
			\hline	
			Cumins & 47.50\% & 55.30\% & 61.31\% \\
			Li    & 53.80\% & 57.80\% & 63.80\% \\
			Wu    & 52.00\% & 57.00\% & 61.39\% \\
			Dai   & 57.20\% & 56.30\% & 63.89\% \\
			Ando  & \multicolumn{1}{c}{\rule[0.05cm]{0.3cm}{0.03mm}} & 53.67\%  & \multicolumn{1}{c}{\rule[0.05cm]{0.3cm}{0.03mm}} \\
			Tri-training &  33.40\% &  42.03\% & 62.41\% \\
			Majority voting  &  59.10\% &  65.00\% & 62.20\% \\
			MCIL  & \textbf{62.20\%} & \textbf{67.00\%} & \textbf{65.47\%} \\
			\hline
	\end{tabular}}%
	\label{tab:comparewithothers}%
	\vspace{-.1in}
\end{table}%

We compared MCIL with the state-of-the-art of SER methods (i.e. Cummins \textit{et al.} \cite{cummins2017image}, Li \textit{et al.} \cite{li2018attention}, Wu \textit{et al.} \cite{wu2019speech}, Dai \textit{et al.} \cite{dai2019learning} and Ando \textit{et al.} \cite{ando2019speech})  to evaluate the effectiveness. Since datasets in those methods are different from ours, to make a fair comparison, we retrain these methods on $D^1$ using their own hyperparameters and learning rates. The final averaged test result is obtained by 5-fold cross-validation. In addition, we implemented tri-training \cite{zhou2005tri} because it also employs more than one classifier to improve the performance. To verify the conclusion of Bahrami's paper \cite{bahrami2010optimally}, the majority voting of five $\{Net^*_i\}$ is used to embody the collective decision-making result. Please note that we choose the DenseNet trained by MCIL, which is the best classifier, to compare with other methods. The results are listed in Table~\ref{tab:comparewithothers}.

On the MAS corpus, MCIL achieves 62.2\% accuracy, which outperforms the other six methods from 3.1\% to 14.7\%. On the IEMOCAP corpus, MCIL also achieves a superior accuracy, outperforms other methods from 2.0\% to 11.7\%. On the FAU-AIBO corpus, MCIL surpasses others from 1.58\% to 4.16\%. The improvement on FAU-AIBO is lower than the ones on the other two databases. The reason is that FAU-AIBO has two categories, which is simpler than multiple categories in MAS and IEMOCAP. This result demonstrates that MCIL is better at handling more complicated ambiguous data.

As Ando’s method \cite{ ando2019speech} required the statistic of experts’ voting to construct the multi-label, it can be only applied to IEMOCAP corpus. MAS does not include the voting information, and experts voted each word instead of the utterance in FAU-AIBO.

Interestingly, tri-training reaches a good result on FAU-AIBO but performs worst on MAS and IEMOCAP. The possible reason might be the number of emotion categories. The votes of three classifiers are not sufficient for the databases with more than two categories. Moreover, tri-training selects more discriminating data and ignore the ambiguous samples during classifiers interacting. Therefore, they perform worse than MCIL.

As for majority voting result, it achieves 59.10\% and 65.00\% accuracy on MAS and IEMOCAP, which only performs slightly worse than MCIL. And on FAU-AIBO, it performs the third worst of all the method. This result verifies the conclusion in Bahrami’s paper, ``groups rarely outperform their best members".

\subsection{Comparison with state-of-the-arts on each emotion}
To evaluate the performance of MCIL and the state-of-the-arts on each emotions, we further investigate the accuracy of those methods on MAS, IEMOCAP, and FAU-AIBO, which are listed in table~\ref{tab:comparisonMAS}, table~\ref{tab:comparisonIEMOCAP} and table~\ref{tab:comparisonFAU}, respectively. As we can observe, these methods perform quite differently on different emotions.

\begin{table}[htbp]
	\centering
	\caption{The comparison with state-of-the-art methods on MAS.}
	\resizebox{88mm}{14.5mm}{
		\begin{tabular}{lcccccc}
			\toprule
			& Angry & Happy & Neutral & Panic & Sadness & Overall\\
			\midrule
			Cumins & 47.00\% & 39.00\% & 59.00\% & 37.50\% & 54.50\% & 47.50\%\\
			Li    & \textbf{82.50\%} & 17.00\% & 66.50\% & 37.50\% & \textbf{65.50\%} & 53.80\% \\
			Wu    & 50.50\% & 40.00\% & 62.00\% & 43.00\% & 64.50\% & 52.00\%\\
			Dai   & 44.50\% & 41.50\% & 70.50\% & \textbf{66.50\%} & 63.00\% & 57.20\%\\
			Tri-training & 26.00\% & 10.00\% & 63.00\% & 28.50\% & 39.50\% & 33.40\%\\
			Majority voting & 60.00\% & 47.00\% & \textbf{79.00\%} & 58.50\% & 51.00\% & 59.10\%\\
			MCIL & \textit{61.50\%} & \textbf{52.50\%} & \textit{78.50\%} & \textit{59.50\%} & 59.00\%  & \textbf{62.20\%} \\
			\bottomrule
	\end{tabular}}%
	\label{tab:comparisonMAS}%
\end{table}%

\begin{table}[htbp]
	\centering
	\caption{The comparison with state-of-the-art methods on IEMOCAP.}
	\resizebox{88mm}{16mm}{
		\begin{tabular}{lccccc}
			\toprule
			& Angry & Happy & Neutral & Sadness & Overall\\
			\midrule
			Cumins & 46.00\% & 52.00\% & 62.00\% & 61.00\% & 55.30\%\\
			Li    & \textbf{78.00\%} & 10.00\% & 67.00\% & 76.00\% & 57.80\% \\
			Wu    & 61.00\% & 46.00\% & 57.00\% & 64.00\% & 57.00\% \\
			Dai   & 63.00\% & 47.00\% & 54.00\% & 61.00\% & 56.30\% \\
			Ando & 46.00\% &	\textbf{58.00\%} & 53.00\% & 56.00\% & 53.67\%\\
			Tri-training & 45.00\% & 24.00\% & 39.00\% & 71.00\% & 42.03\% \\
			Majority voting & 59.00\% & 53.00\% & 68.00\% & \textbf{80.00\%} & 65.00\% \\
			MCIL & \textit{64.00\%} & \textit{52.00\%} & \textbf{74.00\%} & \textit{78.00\%} & \textbf{67.00\%} \\
			\bottomrule
	\end{tabular}}%
	\label{tab:comparisonIEMOCAP}%
\end{table}%

\begin{table}[htbp]
	\centering
	\caption{The comparison with state-of-the-art methods on FAU-AIBO.}
	\begin{tabular}{lccccc}
		\toprule
		& IDL & NEG & Overall\\
		\midrule
		Cumins & 75.87\% & 40.14\% & 61.31\% \\
		Li    & 78.93\% & 41.78\% & 63.80\%\\
		Wu    & 69.22\% & \textbf{50.00\%} & 61.39\% \\
		Dai   & \textbf{91.12\%} & 24.30\% & 63.89\% \\
		Tri-training & 73.28\% & 46.60\% & 62.41\% \\
		Majority voting & 76.51\% & 47.89\% & 62.20\% \\
		MCIL & 77.48\% & \textit{48.00\%} & \textbf{65.47\%}  \\
		\bottomrule
	\end{tabular}%
	\label{tab:comparisonFAU}%
\end{table}%

As shown in table~\ref{tab:comparisonMAS}, on MAS, MCIL is much better than the other five competing methods on the accuracy of the most ambiguous category \textit{happy}. It shows from 5.5\%  to 42.5\% improvement than the state-of-the-art, which demonstrates the effectiveness of interactive learning for ambiguous emotion recognition. As for the less-ambiguous emotion category \textit{neutral},majority voting obtains the best performance, the reason behind this is that all five classifiers performs good and more consistent on \textit{neutral}. And our MCIL achieves the second best accuracy, which obtains at least an 8.0\% improvement, and is only 0.5\% worse than the majority voting result. On the \textit{angry} and \textit{panic} categories, MCIL still ranks at the second, outperforms other four methods. Although, Li's method reaches a remarkable high accuracy on \textit{angry}, but has rather low accuracy on  \textit{happy} and \textit{panic}. In the category of \textit{sadness}, three methods outperform us only from 4.0\% to 6.5\%. Moreover, MCIL method achieves 62.2\% overall accuracy which is a considerable rise from 3.1\% to 14.8\% compared with the other five methods.

On IEMOCAP, as shown in table~\ref{tab:comparisonIEMOCAP}, MCIL achieves better performance on two of the four emotion categories: \textit{neutral} and \textit{sadness}. On \textit{neutral}, our method demonstrates a 7.0\% to 35.0\% raise compared with other methods. A similar trend can be found on \textit{sadness}. As for \textit{happy} and \textit{angry}, MCIL also obtains the second best performance. Again, Li's method performs very good on \textit{angry}, but  the worst on \textit{happy}. For overall accuracy, our MCIL also obtains the considerable increases from 2.0\% to 24.97\% compared with the state-of-the-arts.

Table~\ref{tab:comparisonFAU} shows that these methods perform differently on different emotions on the Fau-AIBO corpus. For example, Dai's method performs better on \textit{IDL}, but the worst on \textit{NEG}. Wu's approach performs better on \textit{NEG}, but achieves the worst accuracy on \textit{IDL}. Both of them indicate that their methods work well for only specific emotions. Contrasts with those methods, the accuracies of two categoris of our MCIL are more balanced. Our method achieves 77.48\% on \textit{IDL}, which is the third best, and 48.0\% on \textit{NEG}, which is the second best. Meanwhile, MCIL gains best overall accuracy, and outperforms other methods from 4.16\% to 1.58\%.

%



\begin{figure}[htb]	
	\setlength{\belowcaptionskip}{-.5cm}
	\begin{center}
		\includegraphics[width=6.5cm]{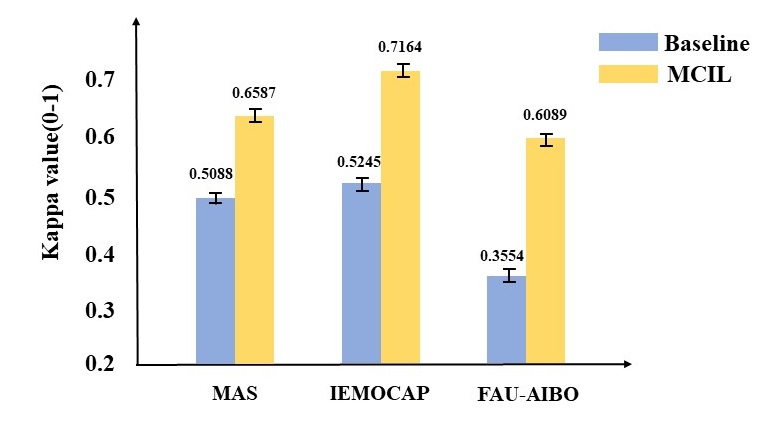}
	\end{center}
	\vspace{-0.2in}
	\caption{The Fleiss Kappa $K$ values of multiple classifiers with and without interactive learning ($p \textless 0.0001$).}
	\label{fig:comparisonforKappavalue}
\end{figure}

\subsection{Consistency evaluation}

The optimally interacting theory also indicates that interpersonal communication can improve cognition consistency among different people. Therefore, we evaluated the classification consistency among five $Net_i$ and $Net_i^*$ using the Fleiss Kappa (\textit{K}) value on three corpora. Figure~\ref{fig:comparisonforKappavalue} illustrates the difference between the \textit{K} values with and without using interactive learning. As we can see, the \textit{K} value of $\{Net_i\}$ trained by the precise labels achieves only 0.5088, 0.5245, and 0.3554 respectively. This is because $D^3$ mainly contains ambiguous samples, which confuse the five $Net_i$ in coming to an agreement. It also shows that $\textit{K}$ rises by 0.15, 0.19, and 0.25 on $\{Net_i^*\}$ after interactive learning, and the consistency has generally been improved from \textit{moderate agreement} to \textit{substantial agreement}.


\section{Conclusion}

In this study, we have addressed the issue of ambiguous SER system by presenting a novel multi-classifier interaction learning (MCIL) method. The MCIL consisted of two novel components: ambiguous label construction and interactive learning. Multi-classifiers were applied to construct ambiguous labels of emotion, which can better represent ambiguous emotion. The interactive learning, which used the KL divergence, was found to be a more feasible strategy for objective measurement. The effectiveness of MCIL was evaluated on three benchmarks: MAS, IEMOCAP, and FAU-AIBO. The experiments show MCIL outperforms state-of-the-art methods on both recognition accuracy and consistency of classification. Both achievements indicated that interactive learning is an effective method for recognizing ambiguous data. Our future study is going to investigate how to improve the robustness of ambiguous label construction.

\appendices

\bibliography{IEEEabrv,sample-sigconf}
\end{document}